\documentclass[review]{cas-sc}

\usepackage[numbers,sort&compress]{natbib}
\usepackage{subfigure}
\usepackage{caption}
\usepackage{amsmath}
\usepackage{hyperref}
\usepackage{stfloats}
\usepackage{wrapfig}
\usepackage{graphicx}
\usepackage{cuted}
\usepackage{lipsum}
\usepackage{svg}
\usepackage{bbding}
\usepackage{setspace}
\usepackage{booktabs}
\usepackage{pifont}

\begin{document}
\doublespacing
\let\WriteBookmarks\relax
\def\floatpagepagefraction{1}
\def\textpagefraction{.001}
\def\sssectionfont{\rmfamily\fontsize{10.5pt}{12pt}%
  \fontseries{b}\selectfont}
\let\printorcid\relax
\captionsetup[figure]{labelfont={bf},labelformat={default},labelsep=period,name={Fig.}}

\shorttitle{}

\shortauthors{Changsong Liu et~al.}

\title [mode = title]{Cycle Pixel Difference Network for Crisp Edge Detection}                      
% \tnotemark[1,2]

% \tnotetext[1]{This document is the results of the research
   % project funded by the National Science Foundation.}

% \tnotetext[2]{The second title footnote which is a longer text matter
%    to fill through the whole text width and overflow into
%    another line in the footnotes area of the first page.}

% \author[1,3]{CV Radhakrishnan}[type=editor,
%                         auid=000,bioid=1,
%                         prefix=Sir,
%                         role=Researcher,
%                         orcid=0000-0001-7511-2910]
\author[1]{Changsong Liu}[style=chinese]

\author[1]{Wei Zhang}[style=chinese]

\author[2]{Yanyan Liu}[style=chinese]
\cormark[1]
\ead[]{lyytianjin@nankai.edu.cn}

\author[1]{Mingyang Li}[style=chinese]

\author[1]{Wenlin Li}[style=chinese]

\author[1]{Yimeng Fan}[style=chinese]

\author[1]{Xiangnan Bai}[style=chinese]

\author[3,4]{Liang Zhang}[style=chinese]

\affiliation[1]{organization={School of Microelectronics, Tianjin University},
    % addressline={Radarweg 29}, 
    city={Tianjin},
    % citysep={}, % Uncomment if no comma needed between city and postcode
    postcode={300072}, 
    country={China}}

% \author[2,4]{Han Theh Thanh}[style=chinese]

% \author[2,3]{CV Rajagopal}[%
%   role=Co-ordinator,
%   suffix=Jr,
% \ead{cvr3@sayahna.org}
% \ead[URL]{www.sayahna.org}

% \credit{Data curation, Writing - Original draft preparation}

\affiliation[2]{organization={College of Electronic Information and Optical Engineering, Nankai University},
    city={Tianjin},
    % citysep={}, % Uncomment if no comma needed between city and postcode
    postcode={300072}, 
    % state={Trivandrum},
    country={China}}

% \affiliation[3]{organization={College of Information Engineering, Luoyang Polytechnic},
%     % addressline={}, 
%     city={Luoyang},
%     % citysep={}, % Uncomment if no comma needed between city and postcode
%     postcode={471000}, 
%     % state={Trivandrum},
%     country={China}}

\affiliation[3]{organization={Tianjin Fire Science and Technology Research Institute of MEM},
    city={Tianjin},
    postcode={300381},
    country={China}}

\affiliation[4]{organization={School of Transportation Science and Engineering, Beihang University},
    city={Beijing},
    postcode={100191},
    country={China}}

% \cormark[1]


% \ead{cvr_1@tug.org.in}

% \ead[url]{www.cvr.cc, cvr@sayahna.org}

% \credit{Conceptualization of this study, Methodology, Software}

% \affiliation[1]{organization={Elsevier B.V.},
%     addressline={Radarweg 29}, 
%     city={Amsterdam},
%     % citysep={}, % Uncomment if no comma needed between city and postcode
%     postcode={1043 NX}, 
%     % state={},
%     country={The Netherlands}}

% % Second author
% \author[2,4]{Han Theh Thanh}[style=chinese]

% % Third author
% \author[2,3]{CV Rajagopal}[%
%    role=Co-ordinator,
%    suffix=Jr,
% \ead{cvr3@sayahna.org}
% \ead[URL]{www.sayahna.org}

% \credit{Data curation, Writing - Original draft preparation}

% % Address/affiliation
% \affiliation[2]{organization={Sayahna Foundation},
%     % addressline={}, 
%     city={Jagathy},
%     % citysep={}, % Uncomment if no comma needed between city and postcode
%     postcode={695014}, 
%     state={Trivandrum},
%     country={India}}

% % Fourth author
% \author%
% [1,3]
% {Rishi T.}
% \cormark[2]
% \fnmark[1,3]
% \ead{rishi@stmdocs.in}
% \ead[URL]{www.stmdocs.in}

% \affiliation[3]{organization={STM Document Engineering Pvt Ltd.},
%     addressline={Mepukada}, 
%     city={Malayinkil},
%     % citysep={}, % Uncomment if no comma needed between city and postcode
%     postcode={695571}, 
%     state={Trivandrum},
%     country={India}}

\cortext[cor1]{Corresponding author}
% \cortext[cor2]{Principal corresponding author}

% \fntext[fn1]{This is the first author footnote. but is common to third
% \fntext[fn2]{Another author footnote, this is a very long footnote and
%   it should be a really long footnote. But this footnote is not yet
%   sufficiently long enough to make two lines of footnote text.}

% \nonumnote{This note has no numbers. In this work we demonstrate $a_b$
%   the formation Y\_1 of a new type of polariton on the interface
%   between a cuprous oxide slab and a polystyrene micro-sphere placed

\begin{abstract}
Edge detection, as a fundamental task in computer vision, has garnered increasing attention. The advent of deep learning has significantly advanced this field. However, recent deep learning-based methods generally face two significant issues: 1) reliance on large-scale pre-trained weights, and 2) generation of thick edges. We construct a U-shape encoder-decoder model named CPD-Net that successfully addresses these two issues simultaneously. In response to issue 1), we propose a novel cycle pixel difference convolution (CPDC), which effectively integrates edge prior knowledge with modern convolution operations, consequently successfully eliminating the dependence on large-scale pre-trained weights. As for issue 2), we construct a multi-scale information enhancement module (MSEM) and a dual residual connection-based (DRC) decoder to enhance the edge location ability of the model, thereby generating crisp and clean contour maps. Comprehensive experiments conducted on four standard benchmarks demonstrate that our method achieves competitive performance on the BSDS500 dataset (ODS=0.813 and AC=0.352), NYUD-V2 (ODS=0.760 and AC=0.223), BIPED dataset (ODS=0.898 and AC=0.426), and CID (ODS=0.59). Our approach provides a novel perspective for addressing these challenges in edge detection.
\end{abstract}

% \includegraphics{figs/grabs.pdf}

% \begin{highlights}
% \item Research highlights item 1
% \item Research highlights item 2
% \item Research highlights item 3
% \end{highlights}

% Keywords

\maketitle

\section{Introduction}
Edge detection is a fundamental task in digital image processing and computer vision that aims to identify points in a digital image where the intensity changes sharply or has discontinuities. These points, commonly organized into a set of curved line segments termed edges, are crucial in understanding image features and content. The significance of edge detection lies in its ability to reduce the amount of data to be processed by filtering out less relevant information while preserving the important structural properties of an image. This process is essential in various applications, including object detection \cite{rani2022object,feng2023boundary,zhang2024advanced,li2024saliency}, image semantic segmentation \cite{9722867,10487946,gu2022net,rehman2018benchmark}, and salient object detection \cite{8288650,10122926,9036909,liu2024two}.

The advent of deep learning has significantly propelled advancements in edge detection. Neural networks, particularly convolutional neural networks (CNNs), have demonstrated remarkable capabilities in learning complex features and patterns \cite{ur2019unsupervised,tu2017csfl,rehman2018optimization}, leading to substantial improvements in edge detection performance and robustness. However, two major issues still need to be addressed currently: (1) most CNN-based methods rely on large-scale pre-trained models, resulting in constraints on the freedom of network architecture design and 
incurring expensive computational costs; (2) most CNN-based methods produce thick and noisy edge maps, which is detrimental to their application in high-level vision tasks.

In this work, we attempt to address these two issues simultaneously. Our goal is to implement training of the network from scratch and guide it to generate crisp and clean edge maps. In response to issue (1), we leverage the idea from traditional gradient-based operators, incorporating such a concept into the modern CNN architecture, developing a cycle pixel difference convolution (CPDC) for targeted and efficient encoding of image edge features. Utilizing the proposed CPDC, we construct fundamental building blocks, upon which we further develop a four-stage backbone network. This backbone network generates features with rich edge prior information, thereby replacing the pre-trained weights on large-scale datasets and conserving computational resources. As for issue (2), the problem of edge thickness reflects the model with insufficient discriminative capability for edge pixels. Therefore, we design a multi-scale information enhancement module (MSEM) that can capture multi-scale contextual information to improve the discriminative ability of the model. We introduce the Squeeze-and-Excitation \cite{hu2018squeeze} channel attention mechanism into MSEM, enabling the model to focus more effectively on edge information. In addition, we build a dual residual connection-based (DRC) decoder to decode deep features into high-quality edge representations, which is crucial for accurate edge detection.

Furthermore, we add a lateral connection between the original input image and the last stage DRC decoder output. The lateral connection includes a ConvNext-V2 \cite{woo2023convnext} module to capture the long-range information for precise object boundary delineation. In the end, we propose a U-shape encoder-decoder network named CPD-Net which consists of these three parts. We perform a series of experiments to demonstrate the effectiveness of CPD-Net, and the main contributions of our work can be summarized as follows:
\begin{enumerate}
    \item We integrate the traditional gradient concept with modern convolution operation, developing a novel cycle pixel difference convolution (CPDC), which enables effective encoding of image edge information, thereby mitigating the reliance on large-scale pre-trained weights.
    \item We propose a U-shape encoder-decoder network named CPD-Net for edge detection. Our CPD-Net mainly consists of three parts: a four-stage backbone based on the CPDC, a multi-scale information enhancement module (MSEM), and a dual residual connection-based (DRC) decoder.
    \item We perform extensive experiments to demonstrate the advantages of CPD-Net and the results show that our method can achieve a competitive performance on four benchmark datasets without any large-scale pre-trained weights.
\end{enumerate}

The structure of this paper is as follows: Section \ref{Related Work} reviews relevant literature in the field. Section \ref{Methodology} provides a detailed description of our method. Section \ref{Experiments} presents a comprehensive evaluation of our method, including the implementation details, the evaluation protocol, an ablation study of each component, and a comparison with current SOTA methods. Finally, Section \ref{Conclusion} concludes the paper with a discussion of our work and identifies its limitations.

\section{Related Work}
\label{Related Work}
Edge detection, a fundamental problem in computer vision, has been the subject of intensive research for over four decades, resulting in a vast body of literature. This section provides an overview of representative works in this field, categorized into two main groups: traditional approaches and deep learning-based methods.

\textbf{Traditional approaches:} Early edge detection approaches primarily relied on image derivative calculations. Roberts \cite{roberts1963machine} introduces a simple first-order derivative operator using diagonal pixel differences, while Sobel \cite{sobel1970camera} employs $3\times3$ kernels for horizontal and vertical gradient computation. Canny \cite{canny1986computational} enhances robustness through a multi-stage process involving noise reduction, gradient calculation, non-maximum suppression, and hysteresis thresholding. The Laplacian detector \cite{jain1995machine} utilizes second-order derivatives to identify rapid intensity changes. Subsequent research integrates additional low-level features such as image texture, color, and gradient information. Methods like Pb \cite{martin2004learning}, gPb \cite{arbelaez2011contour}, and SE \cite{dollar2014fast} employ a classifier to generate object-level boundaries. While these approaches demonstrate improved performance over purely derivative-based methods, their reliance on hand-crafted features and lack of semantic information ultimately constrains their potential for further advancement.

\textbf{Deep learning-based methods:} In recent years, state-of-the-art (SOTA) edge detection methods mainly adopt convolutional neural networks (CNNs). These methods leverage the strong feature extraction ability of CNNs, achieving an impressive performance with higher F-scores, and some of them even surpass humans on several benchmarks. HED \cite{xie2015holistically} employs a fully convolutional neural network with multi-scale side outputs to perform end-to-end edge detection. They also propose a weighted cross-entropy loss function for solving the issue of imbalanced pixel distribution. RCF \cite{liu2017richer} further utilizes features from all convolutional layers in a fully convolutional network to capture both fine details and high-level semantics, resulting in accurate edge maps. BDCN \cite{he2019bi} employs a bi-directional cascade structure to train each network layer with layer-specific supervision, allowing it to focus on edges at different scales. However, these methods generally suffer from the problem of producing excessively thick boundaries. Consequently, researchers have proposed some solutions to address this issue. CED \cite{wang2018deep} introduces a novel refinement architecture for edge detection, incorporating a top-down backward refinement pathway and sub-pixel convolution to improve the location accuracy of edge pixels. LPCB \cite{deng2018learning} employs an encoder-decoder network with a bottom-up/top-down architecture to leverage multi-scale features, enabling the generation of crisp and accurate edge maps without post-processing. DSCD \cite{deng2020deep} introduces a novel loss function based on SSIM \cite{wang2004image} that penalizes the structural difference between prediction and groundtruth, as well as a hyper convolutional module on top of an encoder-decoder network to enhance semantic feature extraction, achieving SOTA performance. DexiNed \cite{soria2023dense} can produce detailed edge maps without any pre-training or fine-tuning process, and they propose a high-quality dataset named BIPED. PiDiNet \cite{su2021pixel} provides a lightweight solution for edge detection by integrating traditional edge detection operators into vanilla convolution in modern DCNN. CATS \cite{huan2021unmixing} proposes a tracing loss that performs feature unmixing by tracing boundaries, and a context-aware fusion block that tackles side mixing by aggregating complementary merits of side edges, resulting in more accurately localized edge predictions without relying on post-processing. EDTER \cite{pu2022edter} utilizes a global transformer encoder to capture long-range context information in Stage I and a local transformer encoder to extract fine-grained cues in Stage II, effectively combining global and local information for accurate edge detection. UAED \cite{zhou2023treasure} models the inherent ambiguity in multiple edge annotations, converting deterministic labels into learnable distributions and leveraging estimated uncertainty to enhance edge detection performance. DiffusionEdge \cite{ye2024diffusionedge} leverages an adaptive FFT filter and an uncertainty distillation strategy for solving the edge thickness.

Although these methods have achieved significant performance, most of them rely on large-scale pre-trained weights, leading to excessive parameters. Therefore, the method proposed in this paper successfully achieves training the network from scratch while simultaneously enhancing the ability of the model to locate edge pixels, thereby generating crisp edge maps.

\section{Methodology}
\label{Methodology}
\begin{figure}[htbp]
    \centering
    \includegraphics[width=\textwidth]{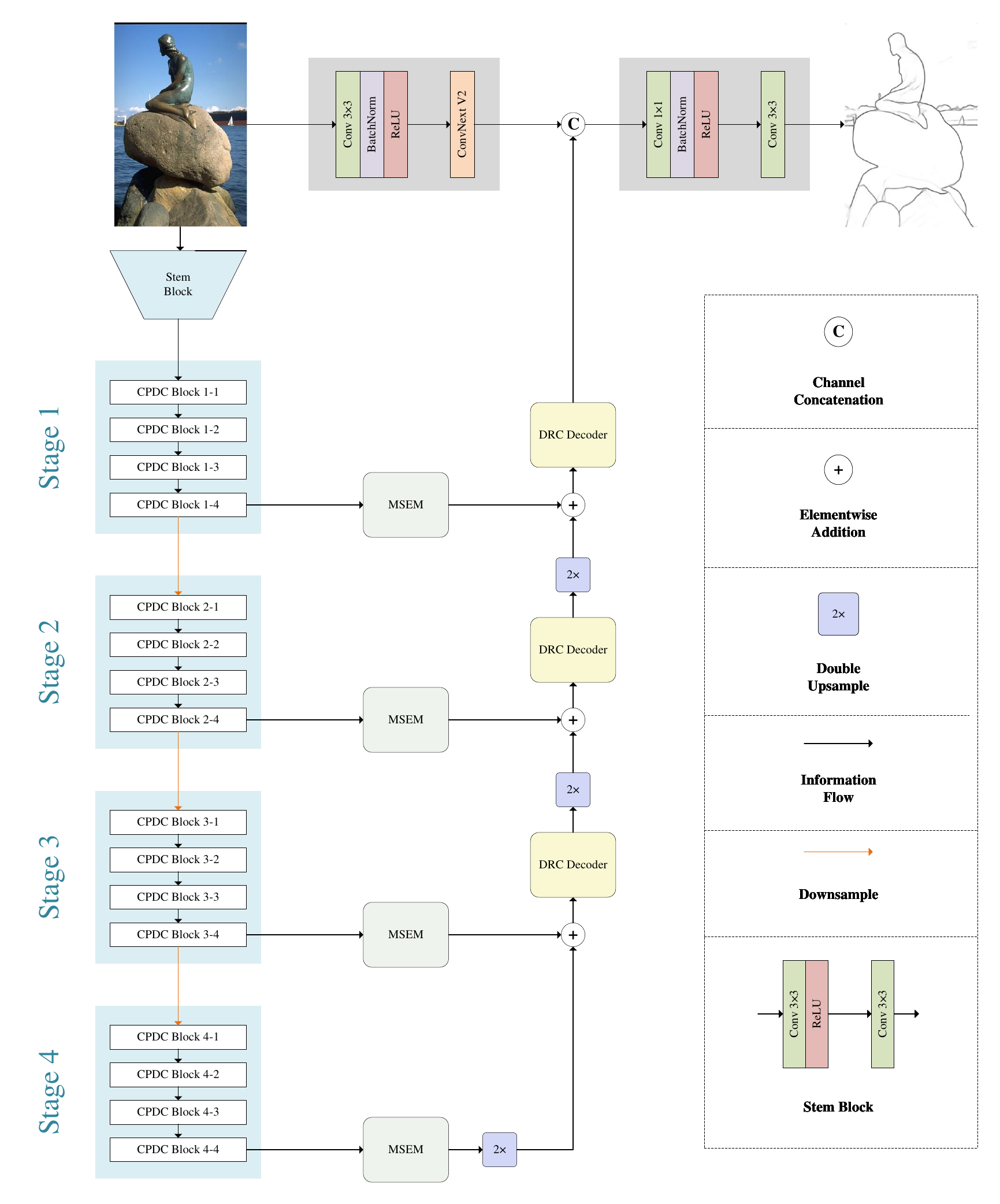}
    \caption{The overall architecture of our proposed cycle pixel difference network (CPD-Net). The whole network adopts a U-shape structure and can be split into four stages. Each stage consists of four CPDC blocks, an MSEM, and a DRC decoder. The MSEM serves as the skip connection to enhance the multi-scale information, and the DRC decoder can decode the features with complete edge information. The final predicted edge maps are obtained by channel-wise concatenation of the outputs from the lateral connection and the DRC decoder.}
    \label{Architecture}
\end{figure}

In this section, we describe our CPD-Net in detail. The whole network adopts an asymmetric U-shape architecture, which mainly consists of three parts as shown in Fig. \ref{Architecture}: encoder, skip-connection, and decoder. Specifically, the encoder part consists of the cycle pixel difference convolution block (CPDC block), the multi-scale information enhancement module (MSEM) serves as the skip-connection, and the decoder component is the dual residual connection-based (DRC) decoder.

Additionally, we add a lateral connection in the whole network and introduce the ConvNext-V2 module into this structure. The ConvNext V2 module employs $7\times7$ convolution, which increases the receptive field, enabling the network to capture a wider range of features. It also utilizes the GELU activation function, effectively mitigating the issue of gradient explosion and enhancing the nonlinear expressive capability of the model. These advantages significantly contribute to improving the completeness of contour extraction. The final prediction of the network is obtained by channel-wise concatenation of the outputs from the lateral connection and the DRC decoder.

\subsection{Cycle pixel difference convolution}
\label{CPDC}
In current SOTA edge detection methods, most of them employ classification networks pre-trained on ImageNet as a backbone, such as VGG \cite{simonyan2014very} and ResNet \cite{he2016deep}, and are then fine-tuned on edge detection datasets through transfer learning to predict the final edge maps. Although such a strategy enhances the performance of edge detection algorithms, it also results in expensive computational costs. Inspired by Sobel operator and Roberts operator, we explicitly encode edge features and embed them into standard convolution, developing four directions of cycle pixel difference convolution (CPDC) operators: CPDC\_v (vertical), CPDC\_h (horizontal), CPDC\_c (cross), and CPDC\_d (diagonal). The proposed CPDC effectively encodes image complete edge features from four specific directions, thereby enabling training the network from scratch and reducing the number of parameters.

Traditional edge detection operators draw contours by computing the gradient of an image, which involves calculating the difference between the values of adjacent pixels. PiDiNet \cite{su2021pixel} follows this idea by integrating image gradient information with convolution operators which are shown in the second row of Fig. \ref{CPDC}. However, both of them only consider the differences between neighbor pixels such as $x_1-x_2$ and $x_1-x_5$, we argue that such a process leads to incomplete edge feature extraction. Therefore, we expand the scope of pixel differences by conducting cycle pixel-wise differencing along four specific directions such as $x_9-x_1$ and $x_7-x_3$, thereby enabling our method to capture both local and long-range spatial information, resulting in preserving more comprehensive edge features. The specific computation process is illustrated in the first row of Fig. \ref{CPDC}.

\begin{figure}[htbp]
    \centering
    \includegraphics[width=\textwidth]{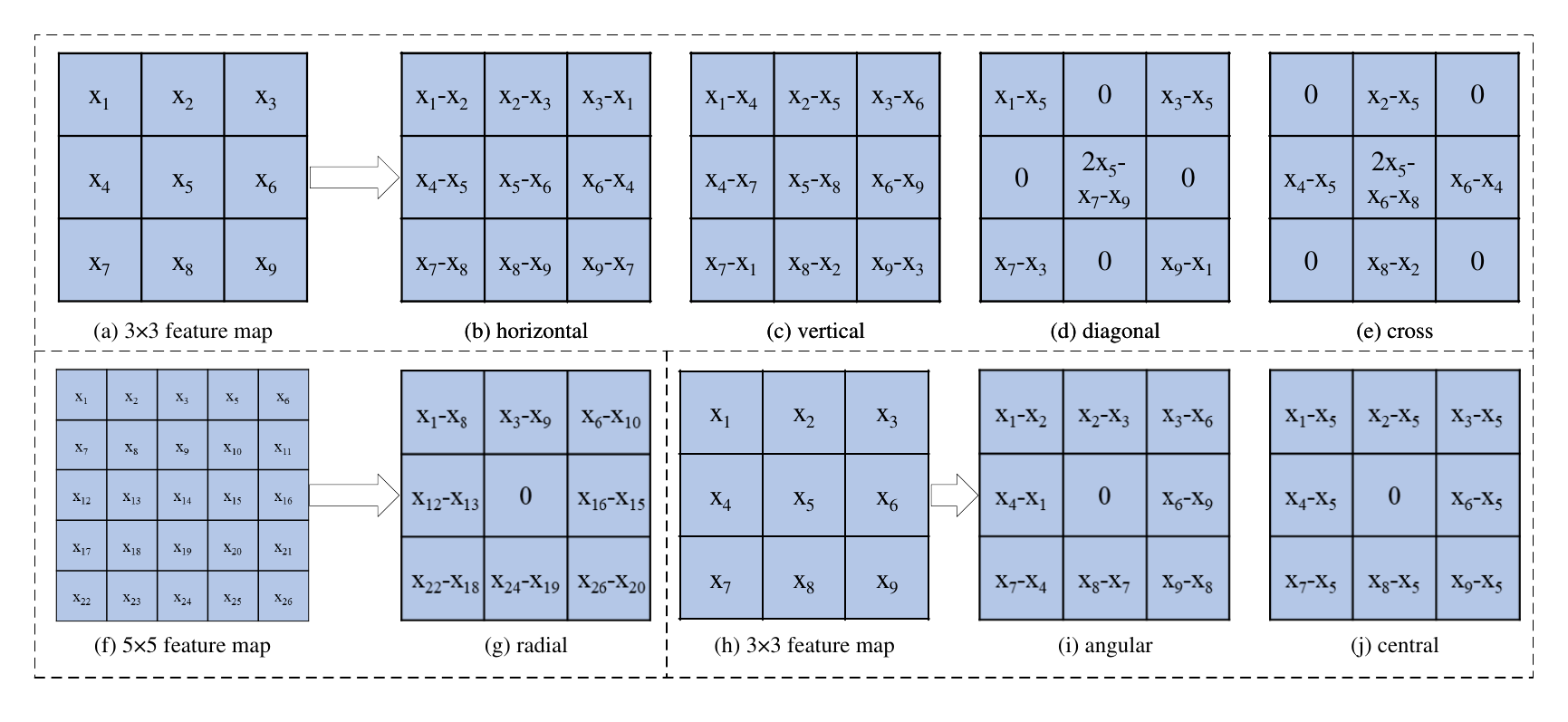}
    \caption{Different types of pixel difference computation. From (a) to (e): (a) indicates a $3\times3$ input feature map, where $x_i$ represents the $i\text{-}th$ pixel value of the feature map. (b) to (e) illustrate the proposed cycle pixel difference operators, which calculate pixel differences in four directions: horizontal, vertical, diagonal, and cross. From (f) to (j): (f) and (h) represent a $5\times5$ and a $3\times3$ feature map, respectively. (g), (i), and (j) represent the pixel difference operators which are proposed in PiDiNet \cite{su2021pixel}.}
    \label{CPDC}
\end{figure}

\begin{figure}[htbp]
	\centering
	\includegraphics[scale=0.6]{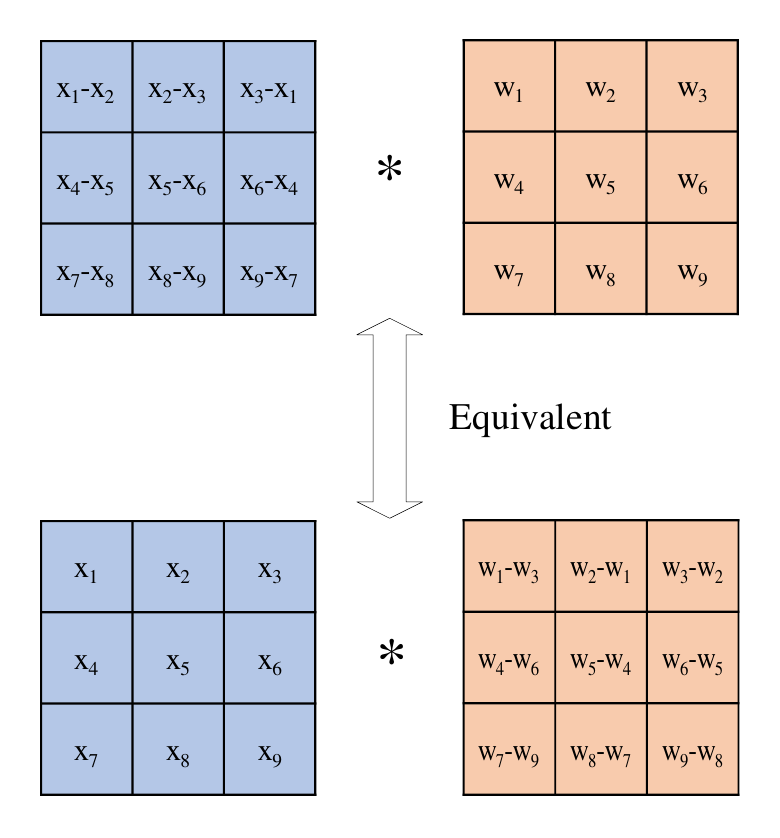}
	\caption{The operation of pixel difference in an image multiplied by a standard convolution weight can be mathematically converted into an equivalent operation where the image is directly multiplied by a weight difference template.}
	\label{conversion}
\end{figure}

Our proposed cycle pixel difference convolution operators can be transformed into a standard convolution with specific weights. Specifically, taking CPDC\_h as an example, given an input image $X$ and a $3\times3$ convolution weight template $W$, the process of CPDC\_h can be written as follows:
\begin{equation}
\begin{aligned}
    f(X;W)&=\left(x_1-x_2\right) \cdot w_1 + \left(x_2-x_3\right) \cdot w_2 + ... + \left(x_9-x_7\right) \cdot w_7\\
          &=x_1 \cdot \left(w_1-w_3\right) + x_2 \cdot \left(w_2-w_1\right) + ... + x_9 \cdot \left(w_9-w_8\right)\\
\end{aligned}
\end{equation}
where $x_i$ indicates the $i\text{-}th$ pixel value in $X$, and $w_i$ indicates the $i\text{-}th$ weight value in $W$. The conversion process is depicted in Fig. \ref{conversion}, and the other three CPDC operators can apply the same transformation. By converting the cycle pixel difference convolution into a standard convolution, we embed complete image gradient information but do not increase additional computational cost, simultaneously leveraging the simplicity and efficiency of traditional gradient-based edge detectors and the powerful feature extraction capabilities of convolutional neural networks, achieving efficient extraction of image edge information.

\begin{figure}[htbp]
	\centering
	\includegraphics[scale=0.4]{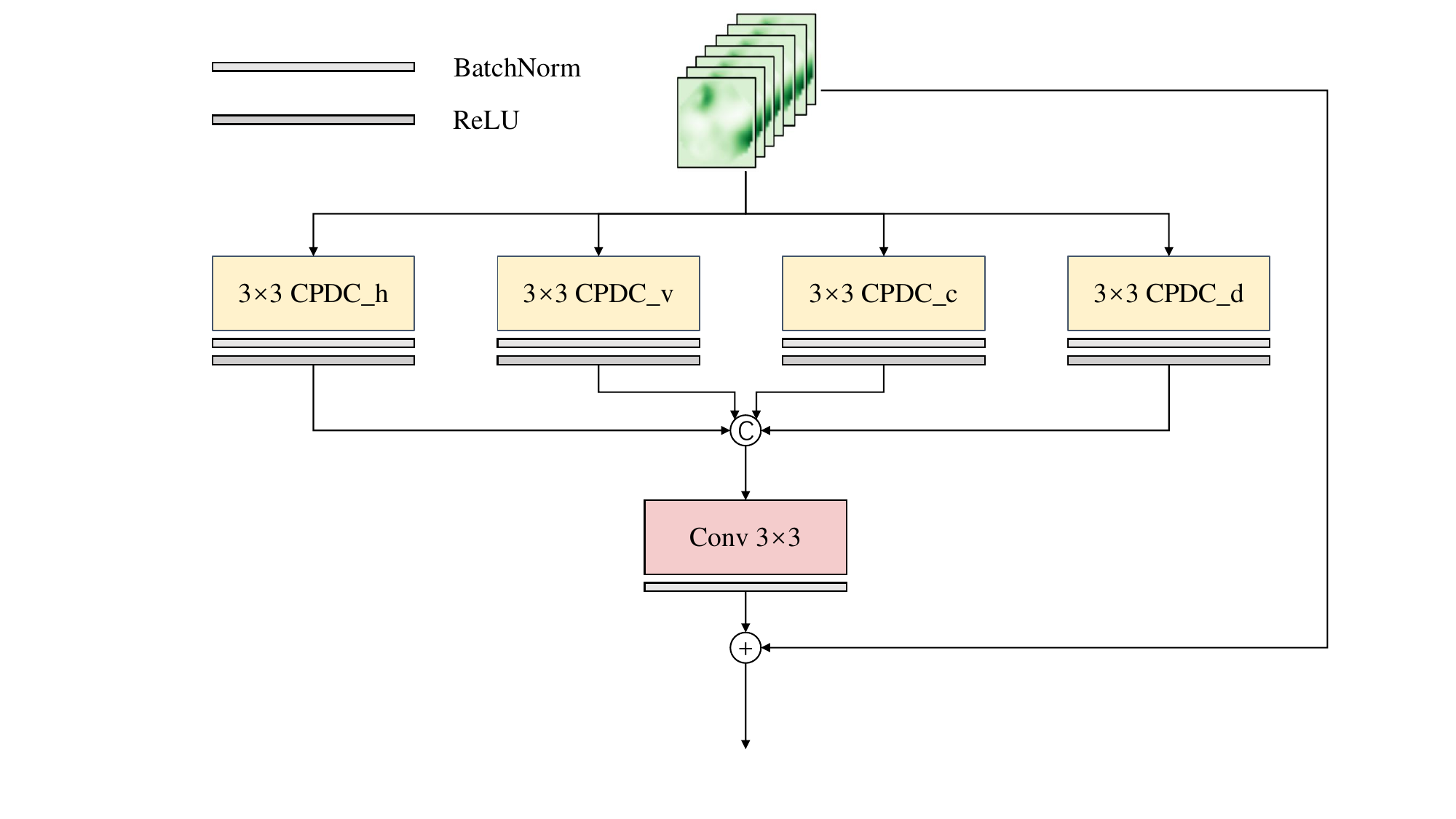}
	\caption{The building block based on cycle pixel difference convolution.}
	\label{CPDC_Block}
\end{figure}

Based on these four forms of cycle pixel difference convolution operators, we develop the CPDC block which can be seen in Fig. \ref{CPDC_Block}. The CPDC block employs the four-branch parallel structure, encoding comprehensive features from four directions. Each branch compresses the channel of the input feature map into $\frac{1}{4}$, then we concatenate the feature maps from the four branches along the channel dimension to restore the original number of channels and use $3\times3$ convolution to integrate the edge information from multi-direction. Additionally, we introduce a skip connection between the input and the output, which can effectively promote information flow and allow the network to learn residual edge information. This architecture leverages multi-directional edge information, leading to more accurate and comprehensive edge detection across different scales and orientations. Finally, we design a four-stage backbone network upon such a building block, with each stage connected by $3\times3$ convolution with the stride of 2, generating feature maps at four different scales. The channel numbers for each stage are set to $\{C, 2C, 4C, 4C\}$, where $C\in\{16, 32, 64\}$ respectively represent Tiny, Small, and Normal versions of our model. Our constructed backbone network is simple yet effective for edge detection, which can eliminate the reliance on large-scale pre-trained weights, and reduce the number of parameters.

\subsection{Multi-scale information enhancement module}
Contemporary CNN-based edge detection methods tend to produce edge maps that are excessively thick and blurry. This quality of edge maps typically reflects the insufficient discriminative capability of the model. Therefore, it is crucial to enhance the discriminative power of the model to locate edge pixels more accurately.

\begin{figure}[htbp]
    \centering
    \includegraphics[scale=0.4]{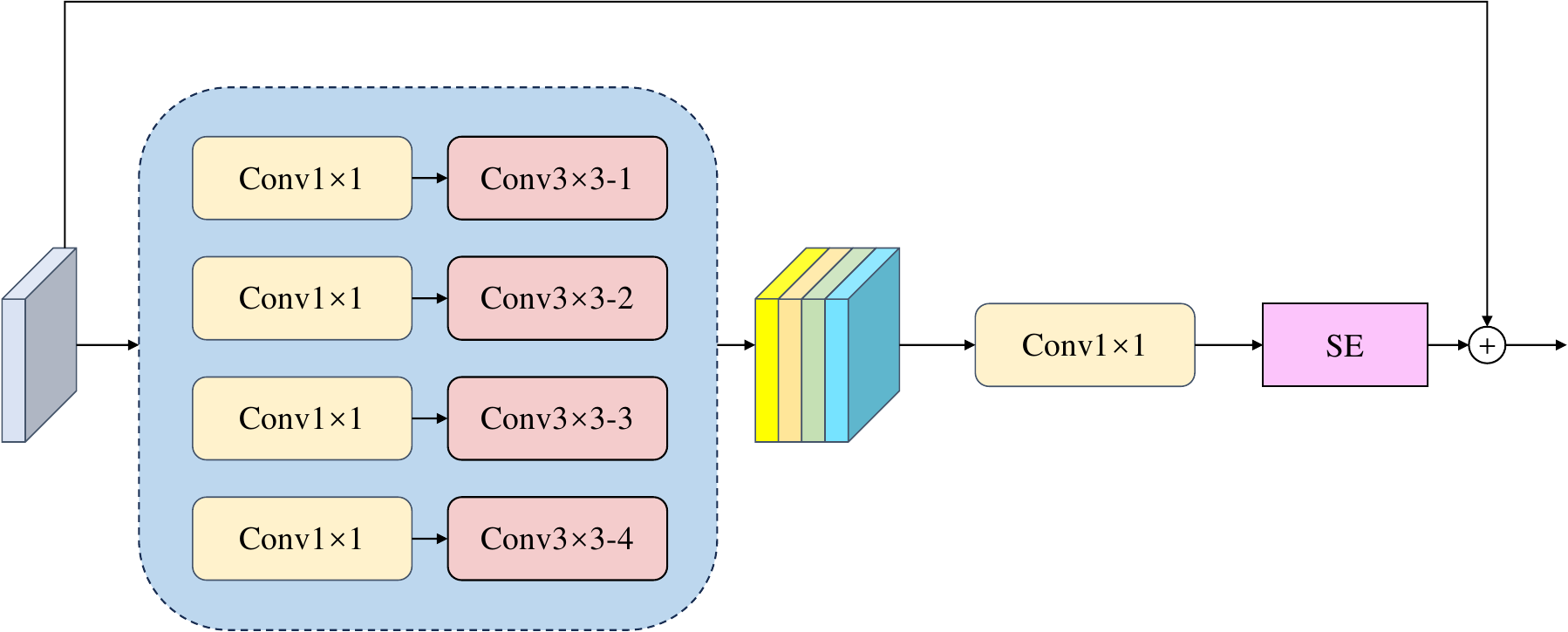}
    \caption{The architecture of MSEM. It consists of four parallel branches, each with $1\times1$ and $3\times3-r$ convolutions where $r$ denotes the dilation rate, followed by channel-wise concatenation, a $1\times1$ convolution, and an SE block. A skip connection links the input to the processed features, allowing the module to adaptively enhance multi-scale information for edge detection.}
    \label{MSEM}
\end{figure}

In edge detection, true edge pixels are typically associated with objects or structures, while noise edges lack semantic coherence. Based on the above analysis, we leverage such characteristics to construct a multi-scale information enhancement module (MSEM). MSEM leverages dilated convolutions with different dilation rates to capture multi-scale information. Smaller dilation rates capture local details, facilitating precise localization of fine edges, while larger dilation rates expand the receptive field, aiding in the capture of long-range information. Therefore, the MSEM can effectively integrate local precise spatial cues from smaller receptive fields with long-range context information from larger receptive fields, resulting in enhancing the network's discriminative ability by providing structural information.

As illustrated in Fig. \ref{MSEM}, MSEM consists of four parallel branches. Each branch initially compresses the feature map channel into 1/4 through $1\times1$ convolution, followed by $3\times3-r$ convolutions with four distinct dilation rates $r\in\{1,2,3,4\}$ to capture multi-scale contextual information. The adoption of four parallel branches is driven by computational efficiency through even channel division. Since the backbone's channel numbers are even at each stage, odd branch numbers (three or five) are impractical, while two branches provide insufficient multi-scale coverage. Therefore, a four-branch structure offers the optimal balance between computational efficiency and multi-scale feature extraction capability. The resulting four scales of feature maps are then concatenated along the channel dimension to restore the number of channels. A $1\times1$ convolution is subsequently applied to the concatenated feature maps for information fusion. Finally, the fused feature maps are passed through the Squeeze-and-Excitation (SE) module \cite{hu2018squeeze}.

The SE attention mechanism enhances channels relevant to edge pixels while suppressing irrelevant ones, implementing an adaptive feature selection mechanism. This strategy increases the signal-to-noise ratio of edge features, thereby enhancing the discriminative capability of the model. Furthermore, we introduce a residual connection between the input and the output, which not only accelerates the network convergence but also promotes the capacity of complex feature representation. The MSEM is formulated as follows:
\begin{equation}
\begin{aligned}
    MSEM = I + SE\left[Conv_{1}\left(\mathop{C}\limits_{r=1}^{4}\sigma\left(Conv_{3}^r\left(\sigma\left(Conv_{1}\left(I\right)\right)\right)\right)\right)\right]
\end{aligned}
\end{equation}
where $I$ denotes the input of MSEM, SE denotes the Squeeze-and-Excitation module, $\mathop{C}$ indicates the channel-wise concatenation, $\sigma$ means the ReLU activation, $r$ is the dilation rate, $Conv$ represents the convolution kernel, and the subscript of $Conv$ means the size of the specific kernel.

By combining multi-scale contextual information with channel attention, MSEM effectively improves the ability of the model to locate edge pixels, resulting in more accurate pixel location. This synergy of multi-scale feature extraction and adaptive feature refinement addresses the limitations of existing methods, producing crisper and more precise edge maps.

\subsection{Dual residual connection-based decoder}
\begin{figure}[htbp]
    \centering
    \includegraphics[width=\textwidth]{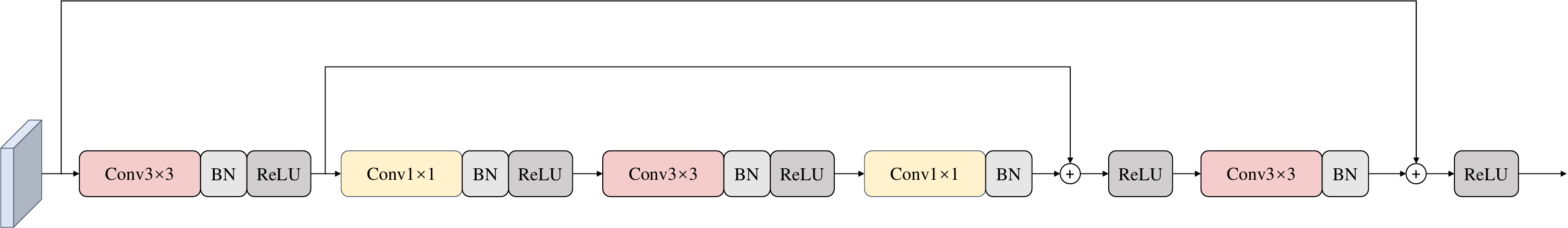}
    \caption{The architecture of DRC decoder. It combines alternating $3\times3$ and $1\times1$ convolutions with BatchNorm and ReLU. The DRC decoder adopts a dual residual structure, created by two skip connections, which enhances feature extraction and gradient flow.}
    \label{DRC}
\end{figure}

The proposed dual residual connection-based (DRC) decoder represents an innovative architecture, which is shown in Fig. \ref{DRC}, to decode and refine edge feature maps. The dual residual design plays a crucial role in enhancing the model's ability to accurately locate edge pixels, primarily by striking a delicate balance between detail preservation and context integration. The first residual connection directly propagates fine-grained information from earlier layers, significantly reducing information loss that typically occurs through several convolutions, thus preserving subtle features critical for precise edge location. Concurrently, the alternating use of $3\times3$ and $1\times1$ convolutions establishes a multi-scale feature extraction mechanism, enabling the model to capture both local details and global semantic information simultaneously. The second residual connection further integrates these multi-scale features, allowing the network to consider both local characteristics and global context in edge determination.

In addition, such a design substantially improves the gradient back-propagation pathway, facilitating effective parameter updates even in deep layers, thereby enhancing the learning and adjustment of features crucial for edge location. Furthermore, this design introduces a flexible adaptive mechanism, enabling the network to dynamically select the most advantageous feature combinations for edge location based on the specific input image characteristics. For instance, in areas with complex textures, the network may rely more heavily on local details, while for large-scale object contours, it might leverage more global information.

In essence, the meticulously crafted DRC decoder enables the network to retain fine details while fully utilizing contextual information, significantly enhancing edge pixel location accuracy through improved learning efficiency and feature representation capacity, thus generating crisp edge maps.

\subsection{Hybrid focal loss function}
We adopt our pioneering work \cite{liu2024learningutilizeimagesecondorder} proposed hybrid focal loss function to address the imbalanced pixel distribution issue in edge detection. The hybrid focal loss function can be decomposed into two constituent parts: the focal Tversky loss and the focal loss. The focal Tversky loss can be written as follows:
\begin{equation}
\begin{aligned}
    L_{FT}=\left(\frac{\sum_{i=1}^{N} p_{i} g_{i}+\left(1-\beta\right) \sum_{i=1}^{N} \left({{p_{i} (1-g_{i})}}\right)^2 +\beta \sum_{i=1}^{N} \left({(1-p_{i}) g_{i}}\right)^2 +C} {\sum_{i=1}^{N} p_{i} g_{i} +C}\right)^\gamma
\end{aligned}
\end{equation}
where $p_i$ and $g_i$ represent the value of $i$-th pixel on an output edge map and its corresponding label image, respectively. $p_{i}g_{i}$, $p_{i}(1-g_{i})$, and $(1-p_{i})g_{i}$ represent true edge pixels (TPs), false edge pixels (FPs), and false non-edge pixels (FNs) in an image, respectively. $\gamma=0.75$ represents the focusing parameter and $C=1\times10^{-7}$ is a constant number to prevent the numerator/denominator from being 0. $1-\beta$ and $\beta$ are the balanced weights between the FPs and FNs. The focal Tversky loss guides the model training process from image-level information.

The focal loss can be defined as $L_{FL}=-\omega\sum_{i=1}^{N}\left[\left(1-p_{i}\right)^\delta g_{i} \log p_{i}+p_{i}^\delta\left(1-g_{i}\right)\log\left(1-p_{i}\right)\right]$ where $N$ represents the total number of pixels in an image. $(1-p_{i})^\delta$ is a modulating factor and $\omega$ is a balance factor for positive and negative pixels. Here, we adopt the original hyper-parameters setting in focal loss, whose $\omega=0.25$ and $\delta=2$. The focal loss constraints the model training process from pixel-level information.

In the end, the whole hybrid focal loss is a weighted fusion between the focal Tversky loss and the focal loss, which can be defined as follows:
\begin{equation}
\begin{aligned}
    L_{H}=L_{FT}+\lambda L_{FL}
\end{aligned}
\end{equation}
where the $\lambda$ to balance the weight between these two parts. The hyper-parameters in focal Tversky loss and $\lambda$ are determined by the ablation experiment. The hybrid focal loss function can effectively address the issue of imbalanced class distribution from both image-level and pixel-level information, consequently guiding the network to generate high-quality edge maps.

\section{Experiments}
\label{Experiments}
In this section, we thoroughly describe the implementation details, including the training hyper-parameters, datasets description, and augmentation strategy. Next, we present the evaluation method we utilize in this study, followed by a series of ablation experiments on our approach. Finally, we compare the CPD-Net with recent SOTA edge detection methods, showcasing the effectiveness of our method.

\subsection{Implementation details}
\label{Implementation details}

\subsubsection{Training hyper-parameters}
\label{Training hyper-parameters}
We use the PyTorch deep learning toolchain to implement our CPD-Net. During the training phase, all datasets are randomly cropped into 320$\times$320 patches and the mini-batch is 8. The initial learning rate is $1\times10^{-4}$, and the number of training epochs is 25. The learning rate is reduced by a factor of 10 every 5 epochs, and the Adam \cite{kingma2014adam} optimizer is employed to optimize the parameters of our network with weight decay $5\times10^{-4}$. For testing, we use a sliding window of 320$\times$320 with stride 240$\times$240, and the predictions in overlapping areas are averaged to produce the final edge maps. We employ the same training parameters across all three versions of CPD-Net, and all the experiments are conducted on a single Tesla A40 GPU.

\subsubsection{Datasets description}
\label{Datasets description}
\textbf{BSDS500:} The Berkeley Segmentation Dataset and Benchmark (BSDS500) is a widely recognized dataset in the field of computer vision, specifically designed for evaluating edge detection algorithms. This dataset comprises 500 natural images, accompanied by multiple human-annotated groundtruth boundaries. The BSDS500 is structured with 200 training images, 100 validation images, and 200 test images. We merge the training images and the validation images into a subset for training, test images for testing, and standard all images to the resolution of $481\times321$. Additionally, following previous works \cite{liu2017richer,deng2018learning}, we augment the BSDS500 dataset with the PASCAL VOC Context dataset, which comprises 10103 images, to further enhance the performance.

\textbf{NYUD-V2:} The NYU Depth Dataset V2 (NYUD-V2) is a significant benchmark dataset in computer vision, particularly valuable for edge detection research. It consists of 1449 RGB-D image pairs with the resolution of $560\times425$ captured from 464 diverse indoor scenes using Microsoft Kinect sensors. The dataset provides rich information for edge detection tasks, including RGB images, corresponding HHA maps, and human-annotated boundary maps. These image pairs are split into 381 for training, 414 for validation, and 654 for testing. We merge the training images and the validation images into a subset for training, test images for testing, and average the predictions from RGB images and HHA images to further improve the performance.

\textbf{BIPED:} The Barcelona Images for Perceptual Edge Detection (BIPED) dataset is a significant contribution to edge detection research, specifically designed to address limitations in existing datasets. Comprising 250 high-resolution ($1280\times720$ pixels) outdoor scenes captured using multiple cameras, BIPED offers diverse environments and lighting conditions. Its key features include pixel-accurate groundtruth edge maps manually annotated by experts, focusing on perceptually relevant edges including both object boundaries and salient texture edges. We adopt their publicly established setting for this dataset, employing a data split comprising 200 images for training and 50 images for testing.

\textbf{CID:} The Contour Image Database (CID) is used for a brain-biologically inspired edge detection method \cite{grigorescu2003contour}. It contains 40 images with natural scenes and each of them is accompanied by an associated groundtruth contour map drawn by a human. The resolution of each image is $512\times512$. The CID exhibits limitations, primarily due to its small scale and the exclusive use of grayscale images. These characteristics pose considerable challenges for CNN-based methods. Therefore, we utilize the BSDS500 dataset for training and all CID images for testing.

\subsubsection{Augmentation strategy}
\label{Augmentation strategy}
To enhance the performance of our model, we employ a data augmentation strategy to expand the datasets. Specifically, our protocol involved a series of geometric transformations applied to the image-label pairs. First, we flip each pair in four directions ($0^{\circ}$, $90^{\circ}$, $180^{\circ}$ and $270^{\circ}$). Subsequently, we rotate each pair at $15^{\circ}$ intervals, encompassing a full $360^{\circ}$ range. Finally, we apply resizing and random cropping operations to these rotated pairs, ensuring that the original resolution is maintained throughout the process. It is noteworthy that this augmentation methodology is consistently applied across all datasets utilized in our study.

\subsection{Evaluation metrics}
\label{Evaluation metrics}
In this work, we calculate three metrics for evaluating the model performance: ODS F-score, OIS F-score, and AP. All of them are based on the Precision $PR=\frac{TP}{TP+FP}$ and Recall $RE=\frac{TP}{TP+FN}$, where $TP$, $FP$, and $FN$ represent the number of correctly classified edge pixels, the number of incorrectly classified edge pixels, and the number of missed edge pixels, respectively. The AP is derived by calculating the area under the Precision-Recall curve. As for F-score($\frac{2\times PR\times RE}{PR+RE}$), it can be calculated in two ways: a) aggregating F-scores across all images, where an optimal fixed threshold has been determined for the whole dataset yields the optimal dataset scale (ODS) F-score; b) aggregating the optimal F-score for each image, which is extracted from all possible confidence thresholds yields the optimal image scale (OIS) F-score. The equations for ODS and OIS are depicted as follows:
\begin{equation}
\begin{split}
    ODS=\max\Big\{\frac{1}{N_{img}}\sum_{i}^{N_{img}}F_{t}^{i}, \forall t\in[0.01,...,0.99]\Big\} \\
    OIS=\frac{1}{N_{img}}\sum_{i}^{N_{img}}\Big\{\max F_{t}^{i}, \forall t\in[0.01,...,0.99]\Big\}
\end{split}
\end{equation}
where $N_{img}$ indicates the total number of images in the dataset, $i$ represents the index of images, $t$ means the confident threshold, and $F$ refers to the F-score. For ODS and OIS, the maximum tolerance distances between predicted edge maps and their corresponding groundtruth images are set to 0.011 for NYUD-V2 and 0.0075 for other datasets.

To comprehensively evaluate the crispness of edge maps, following these previous works \cite{10168686,Ye_Xu_Huang_Yi_Cai_2024}, we also report Standard Evaluation Protocol (S-Eval), Crispness-Emphasized Evaluation Protocol (C-Eval), and Average Crispness (AC). 
S-Eval calculates the F-score after applying post-processing including Non-Maximum Suppression (NMS) and morphological operations to generate thinner edge maps.
C-Eval evaluates predicted edge maps without any post-processing, thereby thinner edge maps typically obtain higher F-scores due to they contain more true positive pixels.
AC quantifies edge crispness as the ratio of post-NMS to pre-NMS pixel value sums, ranging from 0 to 1, with higher values indicating crisper edge maps.

\subsection{Ablation study}
\label{Ablation study}
This subsection presents a series of ablation experiments for evaluating key components and settings of our CPD-Net. All these experiments are conducted on the BSDS500 dataset, due to the similar trends in the changes of ODS, OIS, and AP, we only report the ODS under S-Eval and C-Eval, as well as the AC.

\textbf{Channel number:} We first explore the influence of the channel number in the backbone on overall performance. The comparison results are presented in Table \ref{tab_channel}. As evidenced by the results, there is a clear positive correlation between an increase in channel number and enhanced network performance. It is noteworthy that even our most computationally expensive network configuration maintains a relatively modest parameter size of 9.75M. In the subsequent discussion, we define the models with channel numbers of 16, 32, and 64 as Tiny, Small, and Normal, respectively.

\textbf{Loss function:} We investigate the impact of different $\lambda$ and $\beta$ in hybrid focal loss (HFL) on performance. As presented by rows 1-5 in Table \ref{tab_configuration}, the CPD-Net achieves the best performance at $\lambda=0.001$ and $\beta=0.7$ (ODS(S)=0.775, ODS(C)=0.625, AC=0.291). Therefore, we adopt this setting in HFL. When we replace the loss function with the traditional weighted cross-entropy, rows 6-7 reveal a noticeable decline in model performance (ODS(S)=0.775 VS. ODS(S)=0.766). This comparison demonstrates the superior effectiveness of HFL.

\begin{table}
	\caption{The impact of different channel numbers on performance. ODS(S) and ODS(C) indicate the ODS F-score under the S-Eval and C-Eval, respectively.}
	\label{tab_channel}
	\centering
	\begin{tabular}{c|c|c|c|c}
		\toprule
		Channel number& ODS(S)& ODS(C)& AC& Params\\    
		\hline
        16& 0.775& 0.625& 0.291& 0.63M\\
        32& 0.782& 0.643& 0.303& 2.47M\\
        64& 0.792& 0.660& 0.339& 9.75M\\
	    \bottomrule
	\end{tabular}
\end{table}

\begin{table}
	\caption{Ablation study on network configuration. HFL indicates the hybrid focal loss, and WCE indicates the weighted cross-entropy loss. ODS(S) and ODS(C) indicate the ODS F-score under the S-Eval and C-Eval, respectively. \{h, v, d, c\} represent our CPDC\_h, CPDC\_v, CPDC\_d, CPDC\_c, respectively. \{ad, cd, rd\} represent angular difference, central difference, and radial difference, respectively, which are proposed in PiDiNet \cite{su2021pixel}.}
	\label{tab_configuration}
	\centering
	\begin{tabular}{c|c|c|c|c|ccc|c}
		\toprule
		Method& Stage& Loss& Channel& Operator& ODS(S)& ODS(C)& AC& Params\\
		\hline
        \multirow{22}{*}{\makecell{CPD-Net\\(Tiny)}}& 4& HFL($\lambda=0.1, \beta=0.7$)& \{C,2C,4C,4C\}& \{h,v,d,c\}& 0.772& 0.618& 0.276& 0.63M\\
        & 4& HFL($\lambda=0.01, \beta=0.7$)& \{C,2C,4C,4C\}& \{h,v,d,c\}& 0.773& 0.618& 0.283& 0.63M\\
        & \textbf{4}& \textbf{HFL($\lambda=0.001, \beta=0.7$)}& \textbf{\{C,2C,4C,4C\}}& \textbf{\{h,v,d,c\}}& \textbf{0.775}& \textbf{0.625}& \textbf{0.291}& \textbf{0.63M}\\
        & 4& HFL($\lambda=0.001, \beta=0.8$)& \{C,2C,4C,4C\}& \{h,v,d,c\}& 0.766& 0.611& 0.283& 0.63M\\
        & 4& HFL($\lambda=0.001, \beta=0.6$)& \{C,2C,4C,4C\}& \{h,v,d,c\}& 0.767& 0.612& 0.282& 0.63M\\
        \cline{2-9}
        & 4& WCE& \{C,2C,4C,4C\}& \{h,v,d,c\}& 0.766& 0.505& 0.198& 0.63M\\
        & \textbf{4}& \textbf{HFL($\lambda=0.001, \beta=0.7$)}& \textbf{\{C,2C,4C,4C\}}& \textbf{\{h,v,d,c\}}& \textbf{0.775}& \textbf{0.625}& \textbf{0.291}& \textbf{0.63M}\\     
        \cline{2-9}
        & \textbf{4}& \textbf{HFL($\lambda=0.001, \beta=0.7$)}& \textbf{\{C,2C,4C,4C\}}& \textbf{\{h,v,d,c\}}& \textbf{0.775}& \textbf{0.625}& \textbf{0.291}& \textbf{0.63M}\\
        & 4& HFL($\lambda=0.001, \beta=0.7$)& \{C,2C,4C,8C\}& \{h,v,d,c\}& 0.769& 0.616& 0.283& 1.2M\\
        \cline{2-9}
        & \textbf{4}& \textbf{HFL($\lambda=0.001, \beta=0.7$)}& \textbf{\{C,2C,4C,4C\}}& \textbf{\{h,v,d,c\}}& \textbf{0.775}& \textbf{0.625}& \textbf{0.291}& \textbf{0.63M}\\
        & 5& HFL($\lambda=0.001, \beta=0.7$)& \{C,2C,4C,4C,4C\}& \{h,v,d,c\}& 0.772& 0.616& 0.284& 0.96M\\
        \cline{2-9}
        & 4& HFL($\lambda=0.001, \beta=0.7$)& \{C,2C,4C,4C\}& \{h$\times$4\}& 0.753& 0.604& 0.284& 0.63M\\
        & 4& HFL($\lambda=0.001, \beta=0.7$)& \{C,2C,4C,4C\}& \{v$\times$4\}& 0.753& 0.609& 0.286& 0.63M\\
        & 4& HFL($\lambda=0.001, \beta=0.7$)& \{C,2C,4C,4C\}& \{d$\times$4\}& 0.760& 0.612& 0.285& 0.63M\\
        & 4& HFL($\lambda=0.001, \beta=0.7$)& \{C,2C,4C,4C\}& \{c$\times$4\}& 0.765& 0.610& 0.285& 0.63M\\
        & \textbf{4}& \textbf{HFL($\lambda=0.001, \beta=0.7$)}& \textbf{\{C,2C,4C,4C\}}& \textbf{\{h,v,d,c\}}& \textbf{0.775}& \textbf{0.625}& \textbf{0.291}& \textbf{0.63M}\\
        \cline{2-9}
        & 4& HFL($\lambda=0.001, \beta=0.7$)& \{C,2C,4C,4C\}& \{rd$\times$4\}& 0.741& 0.589& 0.281& 0.63M\\
        & 4& HFL($\lambda=0.001, \beta=0.7$)& \{C,2C,4C,4C\}& \{ad$\times$4\}& 0.747& 0.594& 0.283& 0.63M\\
        & 4& HFL($\lambda=0.001, \beta=0.7$)& \{C,2C,4C,4C\}& \{cd$\times$4\}& 0.752& 0.600& 0.281& 0.63M\\
        & 4& HFL($\lambda=0.001, \beta=0.7$)& \{C,2C,4C,4C\}& \{ad,cd,d,c\}& 0.745& 0.594& 0.280& 0.63M\\
        & 4& HFL($\lambda=0.001, \beta=0.7$)& \{C,2C,4C,4C\}& \{ad,cd,rd,c\}& 0.742& 0.594& 0.280& 0.63M\\
        & \textbf{4}& \textbf{HFL($\lambda=0.001, \beta=0.7$)}& \textbf{\{C,2C,4C,4C\}}& \textbf{\{h,v,d,c\}}& \textbf{0.775}& \textbf{0.625}& \textbf{0.291}& \textbf{0.63M}\\
	    \bottomrule
	\end{tabular}
\end{table}

\begin{table}
	\caption{The ablation study results of each component. SEM denotes the scale enhancement module which is proposed in BDCN \cite{he2019bi}. ODS(S) and ODS(C) indicates the ODS F-score under the S-Eval and C-Eval, respectively.}
	\label{tab_ablation}
	\centering
	\begin{tabular}{c|cccc|ccc|c}
		\toprule
		Method& CPDC Block& MSEM& DRC decoder& SEM& ODS(S)& ODS(C)& AC& Params\\
		\hline
        \multirow{5}{*}{CPD-Net(Tiny)} & \ding{55}& \ding{51}& \ding{51}& \ding{55}& 0.769& 0.618& 0.283& 0.63M\\
        & \ding{51}& \ding{55}& \ding{51}& \ding{55}& 0.773& 0.620& 0.285& 0.59M\\
        & \ding{51}& \ding{51}& \ding{55}& \ding{55}& 0.772& 0.621& 0.288& 0.52M\\
        & \ding{51}& \ding{55}& \ding{51}& \ding{51}& 0.754& 0.608& 0.282& 0.76M\\
        & \ding{51}& \ding{51}& \ding{51}& \ding{55}& \textbf{0.775}& \textbf{0.625}& \textbf{0.291}& 0.63M\\
	    \bottomrule
	\end{tabular}
\end{table}

\textbf{Configurations of backbone:} In Table \ref{tab_configuration}, rows 8-9 and 10-11 respectively reveal performance degradation when increasing the channel multiplier of the final stage from 4 to 8, and when expanding the network from 4 to 5 stages. These two modifications lead to varying degrees of performance decline with more computational cost. We argue that overfitting occurred in both situations, resulting in lower performance.

\textbf{Pixel difference convolution:} In addition, we examine the influence of various pixel difference convolution operators on performance. In Table \ref{tab_configuration}, rows 12-16 demonstrate that utilizing any single type of CPDC in isolation leads to a loss in edge feature encoding, resulting in performance degradation. Conversely, the combination of all four types of CPDC achieves the best performance. We extend our analysis by comparing the pixel difference convolution operators constructed in PiDiNet \cite{su2021pixel} with our proposed method. Rows 17-22 reveal a significant performance deterioration when replacing all operators in the backbone network with PiDiNet's approach. However, substituting PDC\_rd with CPDC\_d results in performance improvement (ODS(S)=742 VS. ODS(S)=0.745). These comparison results provide compelling evidence that our CPDC of computing differences by cycle shifting pixels in multiple directions can encode edge features more comprehensively than PiDiNet's difference computation. This comprehensive encoding consequently achieves an advantage in edge detection.

\textbf{Component:} Finally, we demonstrate the effectiveness of each component, and the results are shown in Table \ref{tab_ablation}. We still adopt the CPD-Net(Tiny) for this ablation study. It can be observed that there is a substantial decline in performance when the cycle pixel difference convolution is replaced with standard convolution. This significant performance gap provides full evidence for the effectiveness of our novel convolution operator. In addition, the removal of the MSEM and the DRC decoder results in varying degrees of performance degradation, which underscores the crucial roles that both the MSEM and DRC decoder play in enhancing the overall capability of the network. Furthermore, we compare our MSEM with the scale enhancement module (SEM) proposed in BDCN \cite{he2019bi}. The experiment results demonstrate that replacing MSEM with SEM leads to significant performance degradation while introducing excessive parameters. This comparison validates the superior multi-scale feature extraction capability of MSEM over SEM.

\subsection{Comparison to SOTA methods}
\label{Comparison to SOTA methods}
In this subsection, we compare our CPD-Net with some other SOTA methods, conducting comparative experiments using the four datasets mentioned earlier: BSDS500 \cite{arbelaez2011contour}, NYUD-V2 \cite{silberman2012indoor}, BIPED \cite{soria2023dense}, and CID \cite{grigorescu2003contour}.

\begin{figure}[htbp]
    \centering
    \includegraphics[width=\textwidth]{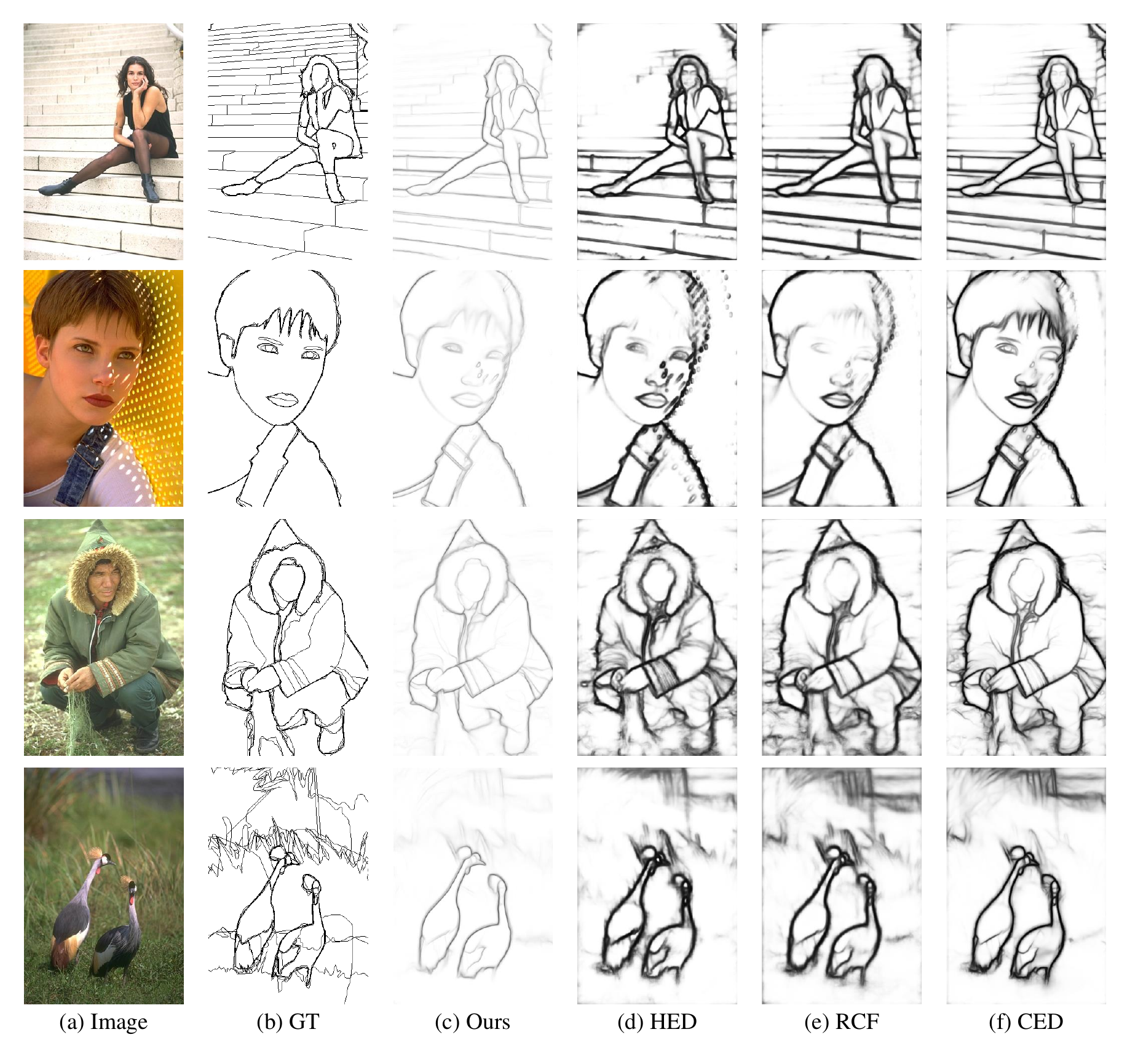}
    \caption{Some examples from different methods. (a) is the original image from the BSDS500 dataset, (b) is its corresponding label image, and (c) to (f) are the predicted edge map from ours, HED, RCF, and CED, respectively. Our method can generate crisp and clean contours.}
    \label{BSDS500_comparison}
\end{figure}

\begin{figure}[htbp]
    \centering
    \includegraphics[width=\textwidth]{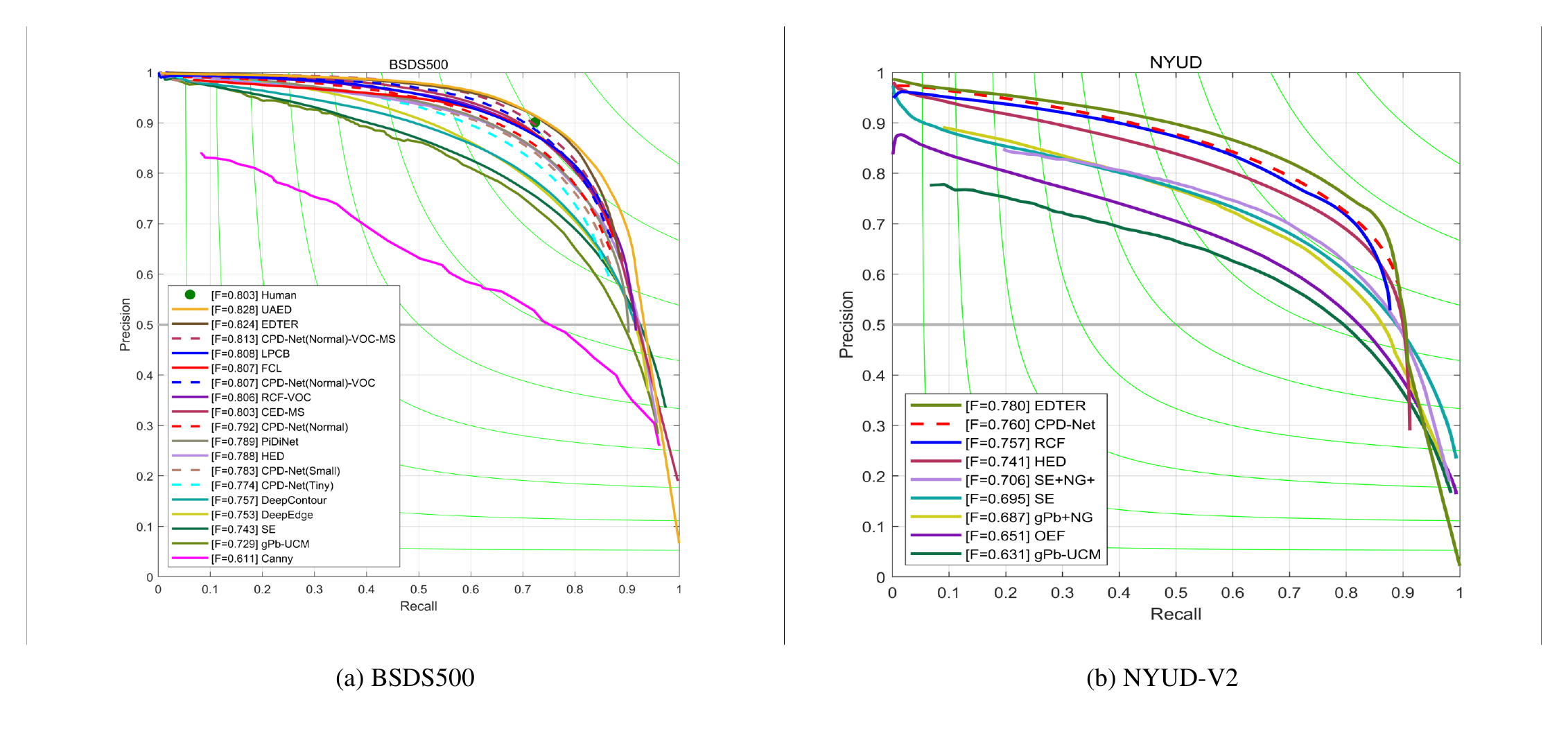}
    \caption{The Precision-Recall curves of BSDS500 and NYUD-V2. These curves provide a comprehensive view of method performance, with curves closer to the top-right corner indicating better performance. Our method achieves a competitive performance on both two datasets.}
    \label{PR_curves}
\end{figure}

\textbf{BSDS500:} Firstly, we evaluate the proposed method against several high-performing edge detection algorithms using the BSDS500 dataset. The comparison includes a selection of recent SOTA edge detectors, which can be categorized into two distinct groups: the first is traditional methods including Canny \cite{canny1986computational}, gPb-UCM \cite{arbelaez2011contour}, and SE \cite{dollar2014fast}; the second is deep learning-based methods including DeepContour \cite{shen2015deepcontour}, DeepEdge \cite{bertasius2015deepedge}, HED \cite{xie2015holistically}, RCF \cite{liu2017richer}, BDCN \cite{he2019bi}, CED \cite{wang2018deep}, LPCB \cite{deng2018learning}, DexiNed \cite{soria2023dense}, PiDiNet \cite{su2021pixel}, CATS \cite{huan2021unmixing}, FCL-Net \cite{XUAN2022248}, EDTER \cite{pu2022edter}, CHRNet \cite{elharrouss2023refined}, PEdger \cite{fu2023practicaledgedetectionrobust}, UAED \cite{zhou2023treasure}, DiffusionEdge \cite{ye2024diffusionedge}, and LUS-Net \cite{liu2024learningutilizeimagesecondorder}. Table \ref{tab_bsds500} presents the quantitative comparison results and Fig. \ref{BSDS500_comparison} shows some examples from different methods. The Precision-Recall curves are drawn in Fig. \ref{PR_curves} (a).

The comparative results presented in Fig. \ref{BSDS500_comparison} demonstrate the advantages of CPD-Net across diverse image types, including portraits, natural scenes, and wildlife. Our approach exhibits remarkable performance in capturing salient edges while preserving fine details, as evidenced by the clear delineation of facial features, clothing textures, and avian plumage. In comparison to established algorithms such as HED, RCF, and CED, our method achieves a superior balance between edge crispness and noise suppression. Notably, it excels in complex textural regions, such as the intricate patterns of foliage and feathers, where other methods over-segment or lose vital details. The results suggest that our method maintains high fidelity to the label image while surpassing existing techniques in terms of edge coherence and detail retention. This comprehensive evaluation across varied visual contexts underscores the robustness and versatility of our proposed edge detection methodology, indicating its potential to advance the SOTA in computer vision applications.

\begin{table}[htbp]
	\caption{Quantitative comparison results on BSDS500 dataset. VOC indicates using extra PASCAL VOC Context dataset in the training process. MS indicates the multi-scale testing. The best and second-best performances are marked in \textcolor{red}{red} and \textcolor{blue}{blue}, respectively. $\ddagger$ indicates the cited GPU speed, and $\dagger$ denotes the inference speed tested on A40 GPU with image resolution of 320$\times$320.}
	\label{tab_bsds500}
	\centering
	\begin{tabular}{c|c|ccc|ccc|c|c|c}
		\toprule
		\multirow{2}{*}{Methods}& \multirow{2}{*}{Pretrained}& \multicolumn{3}{c|}{S-Eval}& \multicolumn{3}{c|}{C-Eval}& \multirow{2}{*}{AC} &\multirow{2}{*}{Params}& \multirow{2}{*}{FPS}\\
        \cline{3-8}
        & & ODS& OIS& AP& ODS& OIS& AP& & \\
		\hline
		Canny \cite{canny1986computational}& -& 0.611& 0.676& 0.520& -& -& -& -& -& -\\
		gPb-UCM \cite{arbelaez2011contour}& -& 0.729& 0.755& 0.745& -& -& -& -& -& -\\
		SE \cite{dollar2014fast}& -& 0.743& 0.764& 0.800& -& -& -& -& -& -\\
		\hline
		DeepEdge \cite{bertasius2015deepedge}& \ding{51}& 0.753& 0.772& 0.807& -& -& -& -& -& -\\
		DeepContour \cite{shen2015deepcontour}& \ding{51}& 0.757& 0.776& 0.790& -& -& -& -& -& -\\
		HED \cite{xie2015holistically}& \ding{51}& 0.788& 0.808& 0.840& 0.588& 0.608& -& 0.215& 14.7M& 30$\ddagger$\\
        HED \cite{xie2015holistically}& \ding{55}& 0.709& 0.724& 0.752& 0.543& 0.567& -& 0.228& 14.7M& 30$\ddagger$\\
		% DRC & 0.802& 0.818& 0.695\\
		CED \cite{wang2018deep}& \ding{51}& 0.794& 0.811& -& 0.642& 0.656& -& 0.207& 14.9M& -\\
        CED-MS \cite{wang2018deep}& \ding{51}& 0.803& 0.820& -& -& -& -& -& 14.9M& -\\
		RCF \cite{liu2017richer}& \ding{51}& 0.798& 0.815& -& 0.585& 0.604& -& 0.189& 14.8M& 30$\ddagger$\\
        RCF \cite{liu2017richer}& \ding{55}& 0.707& 0.726& -& 0.513& 0.516& -& 0.177& 14.8M& 30$\ddagger$\\
		LPCB \cite{deng2018learning}& \ding{51}& 0.800& 0.816& -& 0.693& 0.700& -& -& 15.7M& 30$\ddagger$\\
        BDCN \cite{he2019bi}& \ding{51}& 0.806& 0.826& 0.847& 0.636& 0.650& -& 0.233& 16.3M& 44$\ddagger$\\
        BDCN-VOC-MS \cite{he2019bi}& \ding{51}& 0.828& 0.844& 0.890& -& -& -& -& 16.3M& 44$\ddagger$\\
        PiDiNet \cite{su2021pixel}& \ding{55}& 0.789& 0.803& -& 0.578& 0.587& -& 0.202& \textcolor{blue}{0.71M}& 59$\dagger$\\
        PiDiNet-VOC-MS \cite{su2021pixel}& \ding{55}& 0.807& 0.823& -& 0.602& 0.608& -& 0.201& \textcolor{blue}{0.71M}& 59$\dagger$\\
        DexiNed \cite{soria2023dense}& \ding{55}& 0.729& 0.745& 0.583& -& -& -& -& 35.2M& -\\
        CATS \cite{huan2021unmixing}& \ding{51}& 0.800& 0.816& -& 0.666& 0.676& -& -& -& -\\
        FCL-Net \cite{XUAN2022248}& \ding{51}& 0.807& 0.822& -& -& -& -& -& -& -\\
        EDTER \cite{pu2022edter}& \ding{51}& 0.824& 0.841& 0.880& 0.698& 0.706& -& 0.288& 468.84M& 2.1$\ddagger$\\
        EDTER-MS \cite{pu2022edter}& \ding{51}& \textcolor{red}{0.840}& \textcolor{red}{0.858}& \textcolor{red}{0.896}& -& -& -& -& 468.84M& 2.1$\ddagger$\\
        CHRNet \cite{elharrouss2023refined}& \ding{51}& 0.787& 0.788& 0.801& -& -& -& -& -& -\\
        CHRNet-VOC-MS \cite{elharrouss2023refined}& \ding{51}& 0.816& 0.845& 0.846& -& -& -& -& -& -\\
        PEdger-VOC \cite{fu2023practicaledgedetectionrobust}& \ding{55}& 0.821& 0.841& -& -& -& -& 0.333& 0.73M& 66$\dagger$\\
        UAED \cite{zhou2023treasure}& \ding{51}& 0.829& 0.847& \textcolor{blue}{0.892}& \textcolor{blue}{0.722}& \textcolor{blue}{0.731}& -& 0.227& 72.54M& 25$\dagger$\\
        DiffusionEdge \cite{ye2024diffusionedge}& \ding{51}& \textcolor{blue}{0.834}& \textcolor{blue}{0.848}& -& \textcolor{red}{0.749}& \textcolor{red}{0.754}& -& \textcolor{red}{0.476}& 224.9M& 0.4$\ddagger$\\
        LUS-Net \cite{liu2024learningutilizeimagesecondorder}& \ding{51}& 0.826& 0.846& 0.820& 0.720& 0.726& \textcolor{red}{0.779}& \textcolor{blue}{0.384}& 70.41M& 30$\dagger$\\
		\hline
        CPD-Net(Tiny)& \ding{55}& 0.775& 0.795& 0.783& 0.625& 0.636& 0.657& 0.291& \textcolor{red}{0.63M}& \textcolor{red}{196$\dagger$}\\
        CPD-Net(Small)& \ding{55}& 0.782& 0.803& 0.785& 0.643& 0.651& 0.682& 0.303& 2.47M& \textcolor{blue}{102$\dagger$}\\
        CPD-Net(Normal)& \ding{55}& 0.792& 0.811& 0.798& 0.660& 0.669& 0.707& 0.339& 9.75M& 48$\dagger$\\
        CPD-Net(Normal)-VOC& \ding{55}& 0.807& 0.827& 0.812& 0.687& 0.694& \textcolor{blue}{0.740}& 0.352& 9.75M& 48$\dagger$\\
        CPD-Net(Normal)-VOC-MS& \ding{55}& 0.813& 0.835& 0.840& 0.669& 0.675& 0.723& 0.288& 9.75M& 48$\dagger$\\
	    \bottomrule
	\end{tabular}
\end{table}

The quantitative results shown in Table \ref{tab_bsds500} and Fig. \ref{PR_curves} (a) offer a more direct comparison of various edge detection methods. Remarkably, our method demonstrates highly competitive performance against recent SOTA methods despite not having any pre-trained weights. This is evident in its strong performance across S-Eval, C-Eval and AC. Specifically, CPD-Net(Normal) achieves competitive performance in S-Eval (ODS=0.792, OIS=0.811, AP=0.798,), C-Eval (ODS=0.660, OIS=0.669, AP=0.707) and a high AC of 0.339, surpassing many models with more parameters such as BDCN \cite{he2019bi} and EDTER \cite{pu2022edter}. The VOC dataset further improves these performances, with CPD-Net(Normal)-VOC reaching ODS=0.807 in S-Eval, ODS=0.669 in C-Eval, and AC=0.352. These results are particularly impressive given the model's relatively small size of 9.75M, outperforming CED with 14.9M in crispness-related metrics, which is proposed to solve the thickness problem. Furthermore, CPD-Net(Tiny) achieves remarkable efficiency with 196 FPS while maintaining competitive performance, making it ideal for real-time applications. The highest performance of CPD-Net surpasses that of PiDiNet \cite{su2021pixel} by margins of 3.04\% in ODS for S-Eval, 18.9\% in ODS for C-Eval, and 74.2\% in AC. These comparative results demonstrate that our method effectively mitigates the issue of edge thickness while simultaneously reducing parameters, achieving a balance between accuracy and model complexity. Ultimately, we perform comparative studies to evaluate the performance of HED and RCF with and without pre-trained weights. The comparison results demonstrate that both two methods suffer from substantial performance deterioration without pre-training (10.03\% in HED and 11.4\% in RCF), which proves the prevalent dependency on pre-trained weights among most existing edge detection approaches.

\begin{figure}[htbp]
    \centering
    \includegraphics[width=\textwidth]{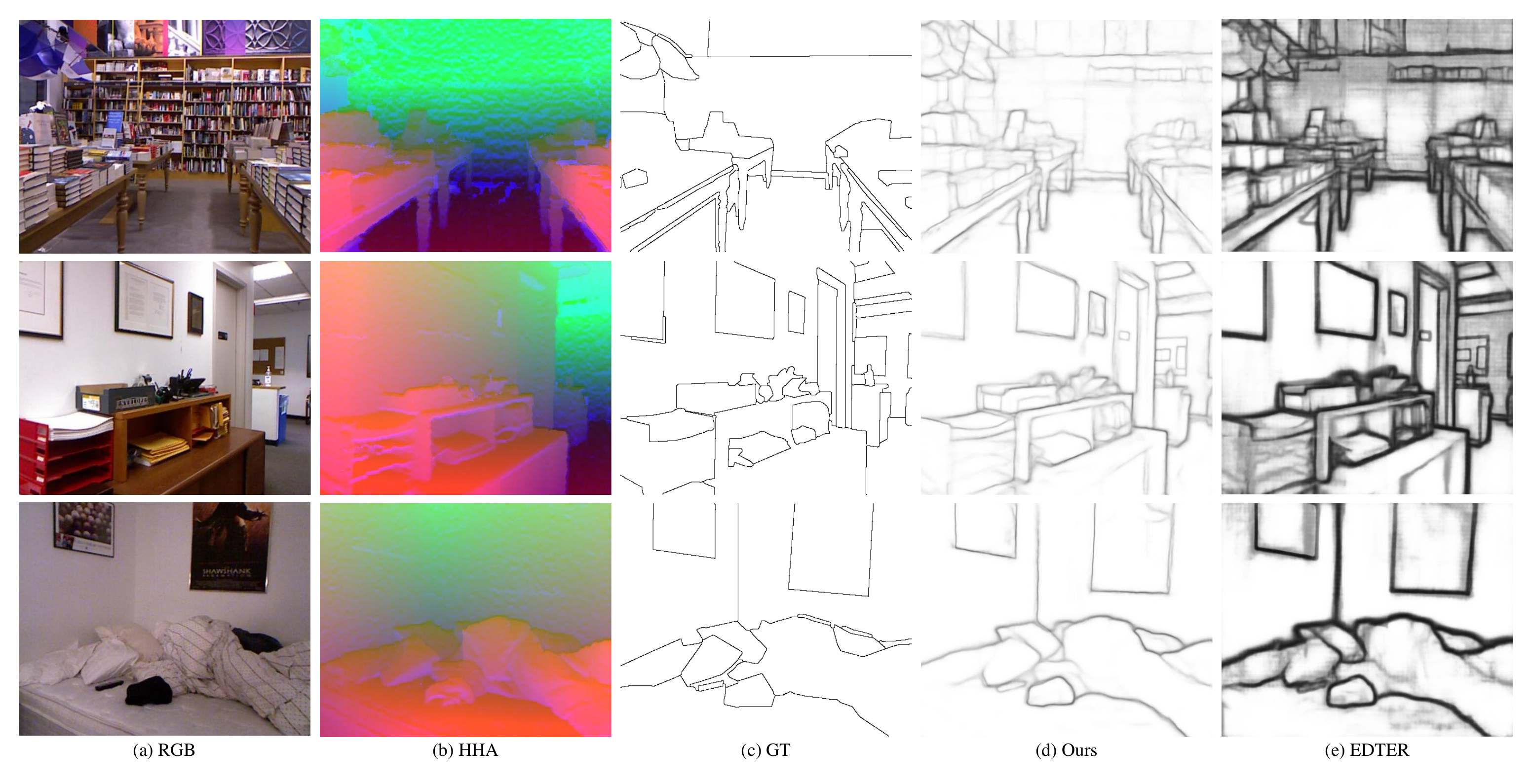}
    \caption{Some examples of different methods on the NYUD-V2 dataset. (a), (b), and (c) are the RGB image, HHA image, and its corresponding label image, respectively. (d) and (e) are the predicted edge maps from our method and EDTER.}
    \label{NYUD}
\end{figure}

\textbf{NYUD-V2:} We perform another comparison experiment on the NYUD-V2 dataset, and we choose recent SOTA methods including gPb-UCM \cite{chen2020contour}, OEF \cite{hallman2015oriented}, gPb+NG \cite{gupta2013perceptual}, SE \cite{dollar2014fast}, SE+NG+ \cite{gupta2014learning}, HED \cite{xie2015holistically}, RCF \cite{liu2017richer}, BDCN \cite{he2019bi}, DexiNed \cite{soria2023dense}, PiDiNet \cite{su2021pixel}, LPCB \cite{deng2018learning}, CHRNet \cite{elharrouss2023refined}, EDTER \cite{pu2022edter}, PEdger \cite{fu2023practicaledgedetectionrobust}, LUS-Net \cite{liu2024learningutilizeimagesecondorder}, and DiffusionEdge \cite{ye2024diffusionedge} for comparison. The qualitative comparison results are shown in Fig. \ref{NYUD}, despite NYUD-V2 predominantly featuring indoor scenes, our method exhibits consistent performance with that observed on the BSDS500 dataset. The contour maps predicted by the CPD-Net are clean and crisp, demonstrating robust generalization capabilities and superior pixel-level discriminative power. Compared with EDTER, which is a powerful edge detection method, our CPD-Net consistently produces more refined and continuous edges that closely align with the groundtruth, suggesting enhanced fidelity to the actual structure scene.

\begin{table}[htbp]
	\caption{Quantitative comparison results on NYUD-V2 dataset. RGB and HHA indicate training with RGB images and HHA images, respectively. RGB-HHA indicates averaging the predicted edge maps from RGB images and HHA images. The best and second-best performances are marked in \textcolor{red}{red} and \textcolor{blue}{blue}, respectively. $\ddagger$ indicates the cited GPU speed, and $\dagger$ denotes the inference speed tested on A40 GPU with image resolution of 320$\times$320. AC is calculated on the RGB images.}
	\label{tab_nyud}
	\centering
	\begin{tabular}{c|ccc|ccc|ccc|c|c|c}
		\toprule
		\multirow{2}{*}{Methods}& \multicolumn{3}{c|}{RGB}& \multicolumn{3}{c|}{HHA}& \multicolumn{3}{c|}{RGB-HHA}& \multirow{2}{*}{AC}& \multirow{2}{*}{Params}& \multirow{2}{*}{FPS}\\
        \cline{2-10}
        &ODS& OIS& AP& ODS& OIS& AP& ODS& OIS& AP& &\\
		\hline
		gPb-UCM \cite{arbelaez2011contour}& 0.631& 0.661& 0.562& -& -& -& -& -& -& -& -& \\
        OEF \cite{hallman2015oriented}& 0.651& 0.667& 0.653& -& -& -& -& -& -& -& -\\
        gPb+NG \cite{gupta2013perceptual}& 0.687& 0.716& 0.629& -& -& -& -& -& -& -& -\\
		SE \cite{dollar2014fast}& 0.695& 0.708& 0.719& -& -& -& -& -& -& -& -\\
        SE+NG+ \cite{gupta2014learning}& 0.706& 0.734& 0.549& -& -& -& -& -& -& -& -\\
		\hline
		HED \cite{xie2015holistically}& 0.720& 0.734& 0.734& 0.682& 0.695& 0.702& 0.746& 0.761& 0.786& -& 14.7M& 20$\ddagger$\\
        RCF \cite{liu2017richer}& 0.729& 0.742& -& 0.705& 0.715& -& 0.757& 0.771& -& -& 14.8M& 20$\ddagger$\\
        BDCN \cite{he2019bi}& 0.748& 0.763& \textcolor{blue}{0.770}& 0.707& 0.719& \textcolor{red}{0.731}& 0.765& 0.781& \textcolor{blue}{0.813}& 0.162& 16.3M& 44$\ddagger$\\
        PiDiNet \cite{su2021pixel}& 0.733& 0.747& -& 0.715& 0.728& -& 0.756& 0.773& -& 0.173& \textcolor{red}{0.71M}& \textcolor{blue}{59$\dagger$}\\
        DexiNed \cite{soria2023dense}& 0.658& 0.674& 0.556& -& -& -& -& -& -& -& 35.2M& -\\
        LPCB \cite{deng2018learning}& 0.739& 0.754& -& 0.707& 0.719& -& 0.762& 0.778& -& -& 15.7M& 30$\ddagger$\\
        CHRNet \cite{elharrouss2023refined}& 0.729& 0.745& -& \textcolor{red}{0.718}& \textcolor{red}{0.731}& -& 0.750& 0.774& -& -& -& -\\
        % RankED & 0.780& 0.793& 0.826& -& -& -& -& -& -\\
        EDTER \cite{pu2022edter}& \textcolor{red}{0.774}& \textcolor{red}{0.789}& \textcolor{red}{0.797}& 0.703& 0.718& \textcolor{blue}{0.727}& \textcolor{red}{0.780}& \textcolor{red}{0.797}& \textcolor{red}{0.814}& 0.195& 468.84M& 2.1$\ddagger$\\
        PEdger \cite{fu2023practicaledgedetectionrobust}& 0.742& 0.757& -& -& -& -& -& -& -& -& \textcolor{blue}{0.73M}& \textcolor{red}{66$\dagger$}\\
        LUS-Net \cite{liu2024learningutilizeimagesecondorder}& 0.757& \textcolor{blue}{0.768}& 0.705& \textcolor{blue}{0.717}& 0.726& 0.644& \textcolor{blue}{0.768}& \textcolor{blue}{0.780}& 0.775& \textcolor{blue}{0.285}& 70.41M& 30$\dagger$\\
        DiffusionEdge \cite{ye2024diffusionedge}& \textcolor{blue}{0.761}& 0.766& -& -& -& -& -& -& -& \textcolor{red}{0.846}& 224.9M& 0.4$\ddagger$\\
		\hline
        CPD-Net& 0.735& 0.750& 0.690& \textcolor{blue}{0.717}& \textcolor{blue}{0.730}& 0.679& 0.760& 0.775& 0.770& 0.223& 9.75M& 48$\dagger$\\
        % Ours-HHA& 0.717& 0.730& 0.679\\
        % Ours-RGB-HHA& \textbf{0.760}& \textbf{0.775}& \textbf{0.770}\\
	    \bottomrule
	\end{tabular}
\end{table}

Table \ref{tab_nyud} and Fig. \ref{PR_curves} (b) present the quantitative results of various edge detection methods on the NYUD-V2 dataset across multiple input modalities (RGB, HHA, and RGB-HHA). While EDTER \cite{pu2022edter} achieves top performance on RGB and RGB-HHA, it does so at the cost of enormous model size (468.84M) and low inference speed (2.1 FPS), limiting its practical applicability. In contrast, CPD-Net emerges as a highly balanced and efficient solution, offering competitive performance across all input modalities (RGB, HHA, and RGB-HHA) while maintaining a relatively compact model size of 9.75M and an impressive inference speed of 48 FPS. The performance of our CPD-Net is particularly on HHA images (ODS=0.717, ODS=0.730), surpassing many larger models including BDCN \cite{he2019bi} and EDTER \cite{pu2022edter}. As for crispness, CPD-Net provides a competitive performance (AC=0.223), surpassing that of EDTER \cite{pu2022edter}, which aligns with the qualitative comparison results. Importantly, CPD-Net strikes an optimal balance between model complexity, speed, and accuracy. It outperforms the models without any pre-trained weights such as PiDiNet \cite{su2021pixel} and DexiNed \cite{soria2023dense} in accuracy while offering comparable speed. It approaches the accuracy of heavier models while maintaining much better efficiency. These comparison results validate the advantages of our approach.

\textbf{BIPED:} We conduct additional comparison experiments utilizing the BIPED dataset and recent top-performing approaches for this comparative analysis: SED \cite{akbarinia2018feedback}, HED \cite{xie2015holistically}, RCF \cite{liu2017richer}, CED-ADM \cite{li2021color}, BDCN \cite{he2019bi}, PiDiNet \cite{su2021pixel}, CATS \cite{huan2021unmixing}, DexiNed \cite{soria2023dense}, LUS-Net \cite{liu2024learningutilizeimagesecondorder}, EDTER \cite{pu2022edter}, and DiffusionEdge \cite{ye2024diffusionedge}. The quantitative results of this comparative study are presented in Table \ref{tab_biped}. Our CPD-Net achieves a top performance among SOTA methods with lower computational cost. Specifically, compared with the recent top edge detector DiffusionEdge \cite{ye2024diffusionedge}, CPD-Net provides higher performance than its (ODS=0.898 VS. 0.899 and OIS=0.903 VS. 0.901), but with fewer parameters (9.75M VS. 224.9M), representing a 23-fold reduction in model size. The AC of our method is higher than PiDiNet \cite{su2021pixel} and DexiNed \cite{soria2023dense}, which demonstrates the effectiveness of our solution. By achieving SOTA performance without relying on pre-trained weights, our method challenges the prevailing paradigm in this field and opens new avenues for developing highly effective, yet computationally efficient, edge detection methods.

\begin{table}[htbp]
	\caption{Quantitative comparison results on BIPED dataset. MS indicates the multi-scale testing. All metrics are calculated under the S-Eval and AC. The best and second-best performances are marked in \textcolor{red}{red} and \textcolor{blue}{blue}, respectively.}
	\label{tab_biped}
	\centering
	\begin{tabular}{c|ccc|c|c}
		\toprule
		Methods& ODS& OIS& AP& AC& Params\\
		\hline
        SED \cite{akbarinia2018feedback}& 0.717& 0.731& 0.756& -& -\\
        HED \cite{xie2015holistically}& 0.829& 0.847& 0.869& -& 14.7M\\
		RCF \cite{liu2017richer}& 0.849& 0.861& 0.906& -& 14.8M\\
        CED-ADM \cite{li2021color}& 0.810& 0.835& 0.869& -& -\\
        BDCN \cite{he2019bi}& 0.890& 0.899& \textcolor{blue}{0.934}& -& 16.3M\\
        PiDiNet \cite{su2021pixel}& 0.868& 0.876& -& 0.232& \textcolor{red}{0.71M}\\
        CATS \cite{huan2021unmixing}& 0.887& 0.892& 0.817& -& -\\
        DexiNed \cite{soria2023dense}& 0.895& 0.900& 0.927& 0.295& 35.2M\\
        % TEED \cite{soria2023tiny}& 0.828& 0.842& -& -\\
        LUS-Net \cite{liu2024learningutilizeimagesecondorder}& \textcolor{red}{0.902}& \textcolor{red}{0.908}& 0.912& \textcolor{blue}{0.464}& 70.41M\\
        EDTER \cite{pu2022edter}& 0.893& 0.898& -& 0.260& 468.84M\\
        DiffusionEdge \cite{ye2024diffusionedge}& \textcolor{blue}{0.899}& 0.901& -& \textcolor{red}{0.849}& 224.9M\\
		\hline
        CPD-Net& 0.896& 0.901& 0.929& 0.426& \textcolor{blue}{9.75M}\\
        CPD-Net-MS& 0.898& \textcolor{blue}{0.903}& \textcolor{red}{0.935}& 0.329& \textcolor{blue}{9.75M}\\
	    \bottomrule
	\end{tabular}
\end{table}

\textbf{CID:} To demonstrate the generalization capability of our method, we test the CPD-Net on CID. We choose the SOTA methods including Canny \cite{canny1986computational}, PC/BC \cite{spratling2012image}, CO \cite{yang2013efficient}, MCI \cite{yang2014multifeature}, SCO \cite{yang2015boundary}, gPb \cite{arbelaez2011contour}, and SED \cite{akbarinia2018feedback} for this comparison. The quantitative results are summarized in Table \ref{tab_cid}. Despite being trained on a different dataset (BSDS500), CPD-Net still achieves highly competitive results on the CID (ODS=0.59, OIS=0.60, AP=0.49), demonstrating its robust generalization capabilities. It is important to acknowledge that the SED \cite{akbarinia2018feedback} achieves the highest performance across all metrics. This discrepancy can be primarily attributed to the lack of prior knowledge for processing grayscale images, as the CPD-Net is trained on RGB images. This mismatch between the training data distribution and the test dataset introduces a challenge.

\begin{table}[htbp]
	\caption{Quantitative comparison results on CID dataset. All metrics are calculated under the S-Eval, and the evaluated results of CPD-Net are trained on BSDS500. The best and second-best performances are marked in \textcolor{red}{red} and \textcolor{blue}{blue}, respectively.}
	\label{tab_cid}
	\centering
	\begin{tabular}{c|ccc}
		\toprule
		Methods& ODS& OIS& AP\\
		\hline
        Canny \cite{canny1986computational}& 0.56& \textcolor{blue}{0.64}& 0.57\\
        PC/BC \cite{spratling2012image}& 0.58& 0.62& 0.42\\
        CO \cite{yang2013efficient}& 0.55& 0.63& 0.57\\
        MCI \cite{yang2014multifeature}& 0.60& 0.63& 0.53\\
        SCO \cite{yang2015boundary}& 0.58& \textcolor{blue}{0.64}& \textcolor{blue}{0.61}\\
        gPb \cite{arbelaez2011contour}& 0.57& 0.61& 0.54\\
        SED \cite{akbarinia2018feedback}& \textcolor{red}{0.65}& \textcolor{red}{0.69}& \textcolor{red}{0.68}\\
		\hline
        CPD-Net& \textcolor{blue}{0.59}& 0.60& 0.49\\
	    \bottomrule
	\end{tabular}
\end{table}

\section{Conclusion}
\label{Conclusion}
% In this work, we proposed a novel cycle pixel difference network (CPD-Net) for crisp edge detection. The key innovations of our approach include: 1) a new backbone based on cycle pixel difference convolution (CPDC) that effectively encodes edge features from four directions, enabling training from scratch without relying on pre-trained weights; 2) a multi-scale information enhancement module (MSEM) that improves the discriminative ability of the model for edge pixels; 3) a dual residual connection-based (DRC) decoder that enhances edge localization. Extensive experiments on four benchmark datasets demonstrate that our method achieves competitive performance compared to SOTA approaches, even without using pre-trained weights. Notably, our approach generates clean and crisp edge maps while maintaining accuracy. This work provides a new perspective on edge detection by showing it is possible to achieve high performance without relying on large-scale pre-trained models, potentially opening new avenues for efficient edge detection in resource-constrained environments.

\textbf{Discussion:} Edge detection has long been dominated by methods that rely heavily on large-scale pre-trained weights. While effective, these approaches have led to two persistent challenges: excessive parameters and inflexible network design. Concurrently, the issue of edge thickness has remained a longstanding problem in this domain. Notably, research addressing both these challenges simultaneously has been relatively scarce. In this work, we propose the CPD-Net, which represents a novel attempt to tackle these dual challenges concurrently. By injecting edge prior knowledge directly into the network architecture, we have developed a model that exhibits sensitivity to edge features without the need for extensive pre-trained weights. Furthermore, the enhancement of multi-scale information through the Multi-Scale Information Enhancement Module (MSEM) bolsters the model's discriminative capabilities, contributing to the alleviation of edge thickness issues. Extensive experimental results across multiple benchmark datasets validate the effectiveness of our approach in mitigating both challenges simultaneously. We posit that addressing these two challenges concurrently is beneficial for advancing the field of edge detection. By reducing reliance on pre-trained weights and improving edge crispness, we open new avenues for efficient and effective edge detection algorithms. This approach is particularly valuable in resource-constrained environments and applications where the use of large-scale pre-trained models may be impractical.

\textbf{Limitation:} a) Despite the improvements achieved by CPD-Net, there remains room for enhancement in the model's ability to precisely localize edge pixels. The current architecture, while effective in many scenarios, still faces challenges in accurately pinpointing edge locations in complex or ambiguous image regions. This limitation underscores the need for further research into more powerful network designs that can boost the model's location capabilities. b) Our experiments revealed limited performance improvements on the CID dataset, highlighting a potential weakness in the model's generalization capabilities across diverse datasets. This observation points to the broader challenge of developing edge detection models that can maintain high performance across varying image distributions, resolutions, and modalities.

\section*{Acknowledgments}
This work is supported by the National Key Research and Development Program of China (No.2022YFC3006302).

% \section{Bibliography}

% \appendix
% \section{My Appendix}
% Appendix sections are coded under \verb+\appendix+.

% \verb+\printcredits+ command is used after appendix sections to list 
% author credit taxonomy contribution roles tagged using \verb+\credit+ 
% in frontmatter.

\printcredits

\bibliographystyle{cas-model2-names}

\bibliography{cas-refs}


% Author biography without author photo.
% Author biography. Author biography. Author biography.
% Author biography. Author biography. Author biography.
% Author biography. Author biography. Author biography.
% Author biography. Author biography. Author biography.
% Author biography. Author biography. Author biography.
% Author biography. Author biography. Author biography.
% Author biography. Author biography. Author biography.
% Author biography. Author biography. Author biography.
% Author biography. Author biography. Author biography.

% \bio{figs/pic1}
% Author biography with author photo.
% Author biography. Author biography. Author biography.
% Author biography. Author biography. Author biography.
% Author biography. Author biography. Author biography.
% Author biography. Author biography. Author biography.
% Author biography. Author biography. Author biography.
% Author biography. Author biography. Author biography.
% Author biography. Author biography. Author biography.
% Author biography. Author biography. Author biography.
% Author biography. Author biography. Author biography.

% \bio{figs/pic1}
% Author biography with author photo.
% Author biography. Author biography. Author biography.
% Author biography. Author biography. Author biography.
% Author biography. Author biography. Author biography.
% Author biography. Author biography. Author biography.

\end{document}